# Amplitude-Based Approach to Evidence Accumulation

Andrew J. Hanson

Computer Science Department
Indiana University
Bloomington, Indiana 47405

**Abstract**

We point out the need to use probability amplitudes rather than probabilities to model evidence accumulation in decision processes involving real physical sensors. Optical information processing systems are given as typical examples of systems that naturally gather evidence in this manner. We derive a new, amplitude-based generalization of the Hough transform technique used for object recognition in machine vision. We argue that one should use *complex* Hough accumulators and *square* their magnitudes to get a proper probabilistic interpretation of the likelihood that an object is present. Finally, we suggest that probability amplitudes may have natural applications in connectionist models, as well as in formulating knowledge-based reasoning problems.

## 1 Introduction — The Need for Amplitudes

Probability computations are the basis of the quantum mechanical theory of measurement, which governs the accumulation of evidence by all physical sensors. However, the particular forms of these probability computations differ from those typically used to treat uncertainty and probabilistic analyses in artificial intelligence. Among the important aspects of quantum probability theory are the following (for an elementary treatment, see, e.g., [8]):

- **Amplitudes.** The fundamental object of quantum calculations is not a probability at all, but a *probability amplitude*, which may be thought of as the square root of a probability; the complex absolute value squared of an amplitude is a probability density.

- **Linear Superposition.** Probability amplitudes typically are solutions to linear differential equations, and therefore combine linearly according to the *superposition principle*.

- **Complex Values.** Probability amplitudes are complex; in particular, they may be negative, so that sums of them combined using the superposition principle may cancel out to give the effect of negative evidence.

- **Uncertainty Principle.** Measurable quantities in quantum mechanics obey the uncertainty principle, which states that there are precise limits on the mutual accuracy of certain simultaneous measurements.

We are thus led to the following conjecture:

*I. Artificial intelligence systems that depend upon evidence accumulated by physical sensors must ultimately treat probability computations in a manner consistent with quantum mechanical measurement principles, and should therefore use probability amplitudes as the basis of their treatment of uncertainty and probability.*

This viewpoint is motivated by the fact that many important physical sensors are in fact acquiring data at the quantum limit — examples include guidance systems that use faint starlight, laser-based systems, and a variety of electromagnetic sensors.

If we extrapolate this first conjecture to include other classes of systems, we are in fact led to an even broader conjecture:

> II. All *probability computations should be based on the calculus of quantum mechanics; in practice, this means that information should be modeled by solutions to wave equations.*

We will examine some suggestive examples that seem to support this viewpoint, but clearly a great deal more work must be done before it is any more than an interesting speculation.

Our purpose in this paper is thus to explore the basic facts of what might be termed the "amplitude-based approach to the accumulation of evidence," and to investigate their possible applications to model-based probabilistic evidence interpretation in artificial intelligence. In particular, we will use the Radon transform as a starting point to derive a new amplitude-based approach to shape recognition that shares many basic characteristics with the Hough transform.

## 2 Overview of amplitude-based probability

The classic example of a problem requiring amplitude-based probability is the computation of optical interference in coherent optics [7]. The problem is solved using the wave model of light to compute patterns of light intensities. If we write the solution of the wave equation for the propagation of monochromatic electromagnetic waves in spacetime $(t, \vec{x})$ as

$$A = a \exp 2\pi i(\omega t - \vec{k} \cdot \vec{x}/c), \quad (1)$$

the measured optical intensity is the complex magnitude squared,

$$I = A^*A = |A|^2. \quad (2)$$

The intensity $I$ is interpreted as a real probability and $A$ is its probability amplitude.

If we have *two* light waves written in the form

$$A_1 = a_1 \exp(-i\phi_1); \ A_2 = a_2 \exp(-i\phi_2)$$

their joint probability (combined optical intensity) is given by the superposition principle to be:

$$\begin{aligned} I_{12} &= |A_1 + A_2|^2 \\ &= |A_1|^2 + |A_2|^2 + A_1 A_2^* + A_1^* A_2 \\ &= I_1 + I_2 + 2\sqrt{I_1 I_2} \cdot \cos(\phi_2 - \phi_1). \end{aligned}$$

Probabilities therefore do not add linearly, while amplitudes do. In particular, if $I_1 = I_2$, and $\phi_1 - \phi_2 = \pi$, the joint probability *vanishes*; this is the interference phenomenon.

Note that it is common for very large numbers of incoherent waves to behave in such a way that their probabilities *do* add:

$$\begin{aligned} I &= \left|\sum_i A_i\right|^2 \\ &= \sum_i |A_i|^2 + 2\sum_{i<j} \sqrt{I_i I_j} \cos(\phi_i - \phi_j) \\ &\approx \sum_i I_i. \end{aligned}$$

The approximation is valid if the interference terms are sufficiently random. This corresponds to a particular classical limit; such behavior for large numbers of objects is typical of quantum mechanics in general.

## 3 Optical pattern recognition

The use of complex probability amplitudes and the principle of superposition of information is not new — it has a long and well-established literature in the field of optical pattern recognition[14, 13, 7]. In this section we briefly review some techniques in the engineering literature on complex optical filters. These methods can be exploited, for example, to locate instances of patterns in images using either actual optical filters or computer simulations of the action of optical filters that are difficult or impossible to realize in practice.

**Optical Filter.** The basic principle of optical filtering is very simple: when an approximately planar wave of monochromatic light shines on a transparency of a photographic image, the image is a real Fourier transform of the image. Shining monochromatic light through the original image, then through the transformed image, and on to a third optical plane produces a spatial correlation transform that is a spot of light at the center of the original image. If instead of using the same images, we use a template of some desired object to produce the transformed second image, and then illuminate that with light from an arbitrary image, spots of light appear in the final optical plane at the location of *each instance* of an area in the original image that is identical in size and orientation to the template.

If $t(x,y)$ represents the intensities of the template image, the result of illuminating the template image with a monochromatic plane wave is essentially a Fourier transform [13],

$$\tilde{t}(k\alpha, k\beta) \approx \iint_{\substack{\text{template} \\ \text{aperture}}} dx\, dy\, t(x,y) \exp(-2\pi i \vec{k} \cdot \vec{x}). \tag{3}$$

Here $\alpha = x_0/z_0$ and $\beta = y_0/z_0$ describe the coordinate $(x_0, y_0, z_0)$ of a (distant) point in the image plane. For example, if the template image is transparent (has $t(x,y) = 1$) in a square of side $a$, the result is a product of *sinc* functions,

$$\tilde{t}(k\alpha, k\beta) = \frac{\sin(\pi a x_0/z_0)}{\pi x_0/z_0} \frac{\sin(\pi a y_0/z_0)}{\pi y_0/z_0}. \tag{4}$$

Shining monochromatic light through a sample image $f(x,y)$, focussing on the filter created by (3), and focussing the result on a final image plane results in spots of light at each position in the sample image that has a similarly oriented square of side $a$. This is the classical optical information processing paradigm for object location[12].

**Relation to Correlation Matching and the Hough Transform.** The optical filter method we have just described is closely related to correlation matching and to the Hough transform, with one crucial difference noted below. First, we recall that the cross-correlation distribution relating a template $t(x,y)$ and a data distribution $f(x,y)$ is defined by

$$\begin{aligned} C(x,y) &= \iint_{\substack{\text{data} \\ \text{plane}}} dx'\, dy'\, f(x+x', y+y')t(x', y') \\ &= \iint_{\substack{\text{data} \\ \text{plane}}} dx'\, dy'\, f(x', y') t(x'-x, y'-y). \end{aligned} \tag{5}$$

(Note that correlation matching is subject to a wide variety of anomalies, e.g., when the data has some random extreme values, so that various techniques are used in practice to normalize $C(x,y)$[19, 3].)

If $f(x,y)$ is a binary-valued function, say, of image edges, and $t(x,y)$ is single point, then we have a situation analogous to a pinhole camera, and $C(x,y) = f(x,y)$ is the image of the data's *probability amplitude*. (Instead of moving the aperture around in front of a plane wave, a true pinhole camera gathers spherical waves at a stationary aperture, thus giving the reversed image $C(-x, -y)$.) If we take $t(x,y) = f(x,y)$, then $C(x,y)$ is what Brown[4] calls the "Feature Point Spread Function," and is identifiable as the autocorrelation function of the template; Brown argues that the standard Hough transform[9, 20], the generalized GHough transform [1], and the CHough variant[4] are all special cases of this formulation. Others (see, e.g., [18]) have also presented arguments on the equivalence of the Hough transform to template matching.

**The Difference.** However, there is an extremely important distinction between the result of an optical transform and a cross-correlation or traditional Hough transform:

> *The convolution in Eq. (3) sums complex probability amplitudes; the photographic emulsion detects only measurable light intensities that are the* square *of the amplitudes. By contrast, cross-correlations and the usual Hough transforms sum only real, amplitude-like quantities, and lack the last step of squaring the result to produce a quantity interpretable as a measured probability.*

That is, without squaring the result, one should not identify a Hough accumulator with a probability. Furthermore, there are a number of standard optical transform procedures that *require* a complex filter in order to achieve optimal extraction of a signal from the data[14], so that *real* template filters of the type used in the standard Hough transform are not in fact sufficient for optical transform applications.

**Families of Amplitude-Based Hough-like Transforms.** For the purposes of this paper, it is sufficient to observe that an entire family of Hough transform variants can be generated by expanding the elementary image data as a vector similar to the terms in a Taylor series expansion:

$$F(x,y) = \{f(x,y), |\vec{\nabla}f(x,y)|, \vec{\nabla}f(x,y), \vec{\nabla}\cdot\vec{\nabla}f(x,y), \ldots\}, (6)$$

where other operations such as binary-valued thresholding may be included if desired.

If we then write a similar expansion $T(x,y)$ for the template, we may understand all Hough transform variants (plus an arbitrarily complex family of new ones) in terms of all the possible combinations of correlations (or, equivalently, optical filters). GHough[1], for example, is obtained by thresholding the vector-vector correlation. Most of the terms with derivatives will include negative contributions, which are interpretable as probability amplitude phase effects; these negative numbers disappear when we square the result to get a true probability such as would be measured by an optical instrument. Combining these negative numbers with additional data can also result in interference phenomena, indicating a practical application of negative correlations; we surmise that, without interference, it is reasonable to interpret negative correlations as positive evidence since a perfect negative correlation corresponds to data that implicitly contains the shape information.

Use of phase information in the template is likely to be extremely important in some cases, so one should not a priori restrict oneself to filters generatable from subimage templates using only real derivative operations such as those in Eq. (6). In particular, the inclusion of complex Fourier filters of various types will supply additional richness to the procedure.

We plan to describe experimental results for these generalized filters elsewhere. Here, we continue with further details of the theoretical issues; we begin by outlining the link between the Hough transform and the Radon transform, and then propose a general amplitude-based transformation formalism for shape recognition tasks.

## 4 The Radon Transform and Parameter Groups of Transformations

The Radon transform[5] provides the unifying link of all the themes treated so far in this paper, as it is a linear, complex transform that is used extensively in the process of reconstruction of data gathered by processes characterized by probability amplitudes rather than probabilities. Furthermore, the classical Radon transform is identical to the Hough transform for straight lines [6]. The Radon transform $\check{f}(r,\vec{\xi})$ of $f(x,y)$ is defined as

$$\check{f}(r,\vec{\xi}) = \iint_{-\infty}^{\infty} dx\,dy\, f(x,y)\delta(r - \vec{\xi}\cdot\vec{x}), \quad (7)$$

where $\vec{\xi}\cdot\vec{\xi} = 1$, so that $\vec{\xi}$ can be identified with the unit vector $(\cos\phi, \sin\phi)$. If we integrate the Radon transform over a one-dimensional Fourier weighting factor,

$$\int dr\, e^{-2\pi i r k}\check{f}(r,\vec{\xi})$$
$$= \iiint dx\,dy\,dr\, f(x,y)e^{-2\pi i r k}\delta(r - \vec{\xi}\cdot\vec{x}), \quad (8)$$

and write the momentum vector in polar coordinates as $\vec{k} = k\vec{\xi}$, we find that the result is just the Fourier transform:

$$\int dr\, e^{-2\pi i r k}\check{f}(r,\vec{\xi})$$
$$= \tilde{f}(k_x = k\cos\phi, k_y = k\sin\phi). \quad (9)$$

Since we can invert the standard Fourier transform on the right hand side of (9) to find the original data $f(x,y)$, knowledge of the Radon transform completely determines $f(x,y)$ as well.

Our treatment extends trivially to $N$ dimensions, where the Radon transform is essentially a Fourier transform over all orientations of a hyperplane in an $N$ dimensional polar coordinate system *omitting* the radial coordinate. The variable $r$ above is identifiable with this radial variable, the perpendicular distance from the origin to the line or hyperplane:

$$x\cos\phi + y\sin\phi = \vec{x}\cdot\vec{\xi} = r. \quad (10)$$

In the form of Eq. (7), we can easily verify [6] that each of a collection of points of the form $f(\vec{x}) = \sum \delta(\vec{x} - \vec{x}_i)$ produces a sinusoidal curve in $(r, \phi)$ space; points in $f(\vec{x})$ lying on the same line produce curves that intersect at a single common point $(r_0, \phi_0)$, the value of the parameters describing the line in the form (10). This Radon transform is thus a precise mathematical formulation of the Hough transform in terms of distributions (see, e.g., [5]).

**Extending the Transformation Group of the Radon Transform.** It is trivial to prove the linearity of the Radon transform — Radon transforms of sums are sums of Radon transforms — and thus to argue that the Radon transform is a natural vehicle to use for the manipulation of probability amplitudes obeying the superposition principle. The same argument holds for Fourier transforms, which are one of the fundamental tools of quantum mechanical analysis. However, unlike the Fourier transform, which is typically understood as the expansion of a signal in representations of the translation group, the Radon transform concept is easily extended to the more general groups of transformations typical of generalized Hough transform accumulators.

If we take a curve $C$ to be parameterized by the coordinate pair $(u(t), v(t))$, then we may write the curve in an arbitrarily translated, scaled, and rotated coordinate system as:

$$\begin{aligned}x(t) &= r\cos\phi + s(u(t)\cos\theta - v(t)\sin\theta)\\ y(t) &= r\sin\phi + s(u(t)\sin\theta + v(t)\cos\theta)\end{aligned} \quad (11)$$

Here we can identify $r\cos\phi$ with the position of the translated origin $x_0$. Thus the Radon transform over the curve becomes

$$\check{f}(C; r, \phi, s, \theta) = \int_C dt \iint dx\,dy$$
$$\delta(x - x(t))\delta(y - y(t))f(x,y), \quad (12)$$

where $x(t), y(t)$ are given by Eq. (11). Since the variables appearing in the transform are the parameters of a general *group of linear coordinate transformations*, we will simplify our notation by representing the transformation (11) in terms of these group parameters by the symbol $G$; that is,

$$\begin{aligned}G(u,v) &= \\ G(r, \phi, s, \theta; u, v) &= G(\vec{x}_0, s, \theta; u, v)\\ &= (x(t), y(t)).\end{aligned}$$

The generalization to $N$ dimensions is obvious, with $\theta$ being replaced effectively by matrices $\|R\|$ of the $N$ dimensional rotation group. $G$

has 4, 7, and 11 parameters in 2, 3, and 4 dimensions, respectively. There may be circumstances in which conformal transformations[1] as well as other transformations such as skewing and perspective projection should be included in $G$; this seems an elegant way to gather evidence concerning the viewing parameters of a three-dimensional scene, for example.

Thus we may write our extended Radon transform as

$$\check{f}(C;G) = \int_C dt \int d^N x\, f(\vec{x}) \delta^N(\vec{x} - \vec{x}(t)) \quad (13)$$

where

$$\vec{x}(t) = \vec{x}_0 + s\|R\| \cdot \vec{u}(t).$$

However, our formalism is still restricted to single curves representing edge shape information. In the next section, we present the extensions needed to incorporate the full richness of both the image and the template data sources.

## 5 A New Calculus for Probabilistic Shape Recognition

The Radon transform in the form (13) refers to its shape template only implicitly by means of the binary-valued image corresponding to the integration path $C$. This is unsatisfactory if we wish to incorporate other kinds of template paradigms.

We therefore introduce an explicit template structure $T(x, y)$ that is assumed to correspond to image gray-scale data or some analogous quantity that is represented in the same language as a body of evidence $F(x, y)$, expanded as in Eq. (6). Then we choose a suitable group $G$ of coordinate transformations, which in two dimensions would typically be

$$G(x, y) = (x_0 + s(x\cos\theta - y\sin\theta),$$
$$y_0 + s(x\sin\theta + y\cos\theta)),$$

---

[1]Inversions about an arbitrary origin that combine with scaling, translation, and rotations to form the conformal group, with 6, 10, and 15 parameters in dimensions 2,3, and 4.

and define a family of probability amplitudes of the form

$$\begin{aligned} A_{ij}(T;G) &= \iint dx\, dy\, T_i(x,y) F_j(G(x,y)) \\ &= \iint |J|\, dx\, dy\, T_i(G^{-1}(x,y)) F_j(x,y), \end{aligned} \quad (14)$$

where $|J| = s^{-2}$ is the Jacobian of the coordinate transformation. Note that Eq. (14) is essentially an optical filter equation, so there is no reason for either the template or the data to be real numbers; $A_{ij}$ is a probability amplitude, and can in general be complex. In fact, complex filters are widely used in optics applications[13].

Our most basic transformation is then simply the generalization of the cross correlation to include an arbitrary group $G$ of transformations:

$$\begin{aligned} A_{00}(T;G) &= \iint dx\, dy\, t(x,y) f(G(x,y)) \\ &= \iint |J|\, dx\, dy\, t(G^{-1}(x,y)) f(x,y). \end{aligned} \quad (15)$$

For pure translations, we have $t(G^{-1}(x,y)) = t(x - x_0, y - y_0)$ and so we find the usual template correlation filter. Examples of other filters appearing in the expansion of Eq. (6) include

$$\begin{aligned} A_{11}(T;G) &= \iint dx\, dy\, |\vec{\nabla} t(x,y)| |\vec{\nabla} f(G(x,y))| \\ A_{22}(T;G) &= \iint dx\, dy\, \vec{\nabla} t(x,y) \cdot \vec{\nabla} f(G(x,y)) \\ A_{33}(T;G) &= \iint dx\, dy\, \nabla^2 t(x,y) \nabla^2 f(G(x,y)). \end{aligned}$$

Specific variants like binary edge correlation and GHough are found by inserting threshold filters into $A_{11}$ and $A_{22}$.

The Radon transform Eq. (7) is given by the special case

$$A_{\text{Radon}}(T;G) = A_{10}(T;G) = \iint dx\, dy\, |\vec{\nabla} t(x,y)|\, f(G(x,y)), \quad (16)$$

where we take $t(x,y)$ to be the step edge at $x = 0$, so $|\vec{\nabla} t(x,y)| = \delta(x)$. Then if we set the scale to unity, we find

$$\iint dx\, dy\, \delta(x)$$
$$f(x_0 + x\cos\theta - y\sin\theta, y_0 + x\sin\theta + y\cos\theta)$$
$$= \iint dx'\, dy'\, \delta(x'\cos\theta + y'\sin\theta - r)f(x',y'),$$

where we used $(x_0, y_0) = (r\cos\theta, r\sin\theta)$ to make $x_0\cos\theta + y_0\sin\theta = r$. (Because of the translation symmetry of a straight line, the apparent pair of parameters $(x_0, y_0)$ determines uniquely only a single parameter, the distance $r$ of the nearest approach of the line to the origin; we are therefore free to replace the general value $x_0 = r\cos\phi$ by $x_0 = r\cos\theta$.) Thus we have proven that a parameter-free template in our formalism produces exactly the Radon transform. Alternatively, one could set $G$ to the identity, and use a parameterized template in $t(x,y)$. Similar arguments lead to the the Hough transform for a circular template. (The symmetry causes the $\theta$ dependence to disappear, just as $\phi$ dependence disappeared for the straight line).

Many other filters could obviously be similarly defined, and the generalization to three-dimensional data can be carried out following Ballard[2].

**Complex Superposition of Edge and Area Information.** The general form of the probability amplitude for the occurrence of a shape template given the evidence $F(x,y)$ is then written using the superposition principle as

$$A(T;G) = \sum_k c_k A_k(T;G), \qquad (17)$$

where $k$ ranges over all the filters one wishes to apply. This amplitude can be used to combine the evidence for *area* and *edge* information in any way that is semantically meaningful using the complex coefficients $\{c_k\}$. The determination of the coefficients is a semantic problem that depends on the balance attributed to the different evidence sources in the overall model; negative terms, for example, could be used to induce interference effects. In addition, a term in the probability amplitude for the a priori occurrence of particular *geometric shapes* in the two or three-dimensional data could in principle be incorporated to handle generic or constrained shape models[10].

## 6 Probability Amplitudes, Connection Networks, and Semantic Wave Equations

Having dealt mainly with pattern recognition problems in the main body of the paper, we now present some speculations on the possible applications of the probability amplitude approach to analyzing artificial intelligence problems in other domains.

**Connection Networks.** The classic application of probability amplitudes to problem solving is in optical filtering[13, 7, 14]. In these domains, a signal, usually a beam of coherent light focused on a photographic transparency, is transformed to achieve some desired effect by passing it through any number of complex filters. The result can be a deblurred image or an image with bright dots corresponding to image areas having high correlations with a template. Note that there are also electronic circuit analogies to such transformations[13]. The propagation and filtering procedures in optical filtering bear a remarkable resemblance to the propagation of a signal through a connection network [17]: all computations are parallel, and data propagating from one point at one layer can combine with other propagating data at distant points in another layer to cause complicated effects. The difference is that complex probability amplitudes *combine in a much different way* than the data streams currently used in connection networks. It would be very interesting to explore the potential of using complex quantities. Furthermore, there is a concept of "semantic layering" that is present in the optical filtering applications that could be of use in connectionist approaches: while the transformation undergone in each step of the filter is potentially difficult to understand, it will often produce an interpretable intermediate result that is passed on to the next layer of filters. One can therefore envision connection networks built in semantic layers, with each intervening filter making a transformation from either a data representation or a knowledge-space representation to a new "meaningful" knowledge space until reaching the final desired space used for making a decision. Debugging opportunities and explanations of the "procedures" are then embodied in the intermediate layers of results, which can be intercepted and examined.

**Semantic Wave Equations.** We began this paper by arguing that amplitude-based computations were essential for systems that had to base their analysis on the output of real world measuring instruments, since such measurement processes are all phrasable in terms of the quantum mechanical theory of measurement. However, there is no particular reason to believe that artificially intelligent systems, whose goal is to mimic the action of human cognitive abilities, would be efficiently expressed in quantum mechanical terms. Nevertheless, it is exciting to imagine the possibility that there is some inherent elegance achieved by using probability amplitudes for cognitive processes, as well as quantum measurement, that remains to be discovered.

We note in particular that there are many procedures now carried out by computer simulation that have *no analog in nature*. An example is the generalized Hough transform, which can find not only the location, but also the scale and rotation parameters of the shape — no single optical filter can do this, but a computer simulation can. Might it not be the case that there are wave equations describing domains that do not occur in physical reality, but that simulate cognitive processes? There is no physical quantum theory for the density of probabilities of a semantic proposition, but there could well be *wave equations of knowledge propagation* that would model such questions extremely well: the result of solving such an equation would be a probability amplitude for the proposition, and this could be combined using the superposition principle with other waves to generate a total probabilistic analysis of a problem involving many interacting pieces of evidence and semantic knowledge. Furthermore, certain types of analysis that have no clear theoretical foundation at present might become tractable: in particular, it might be possible to determine an "uncertainty principle" limiting the precision to which different bodies of knowledge could be simultaneously determined, and to precisely quantify issues such as "resolving power" to place limits on the reliabilities of conclusions in the presence of uncertainty and conflicting evidence.

# 7 Conclusion

We have proposed an amplitude-based formalism for evidence accumulation in artificial intelligence problems, which has the distinguishing feature that probability amplitudes, not probabilities, are the fundamental computational element. Probability amplitudes superimpose linearly, and can result in complex interference patterns in their squares, which are identified as probabilities. Such an approach is the only one consistent with modeling data from real physical sensors.

As an explicit example of the proposed mechanisms, we have explored optical transform theory, related it to the Hough transform and the Radon transform, and deduced a new and more general object recognition transformation. The extended transform has the potential for incorporating many different aspects of the object recognition problem in ways that have clearly correct probabilistic interpretations.

Among the questions that remain to be investigated more thoroughly are the following:

- **Object recognition.** The structures we have presented essentially unify all Hough-like object recognition techniques. Much

remains to be done to see whether these concepts can be effectively exploited, e.g., to combine edge and area-based evidence with geometric likelihoods to generate reliable probabilistic shape identifications.

- **Invertibility, stereography, and viewpoint reconstruction.** The Radon transform, because of its rigorous invertibility, has found extensive use in spatial data reconstruction for dense data such as CT scans[5, 19]. The invertibility properties of the transformations represented by Eq. (14) are much less clear; their exact inversion characteristics, and the limitations on digital approximations to inversion techniques with sparse data should be understood more completely. Since multiple-view stereographic data has properties similar to CT scans, these transforms can presumably be used with some degree of accuracy to carry out surface reconstruction and the determination of viewpoint parameters. The combined transformations proposed here may be able to achieve an improved unification of edge-based and area-correlation techniques in these reconstruction problems.

- **Relation to Minimal Description Length principle.** Fua and Hanson[10] have extensively explored another probabilistic method, the Minimal Description Length principle[15, 16], to incorporate edge, area, and geometric factors into the object recognition problem. The relationship between the two philosophies is yet to be analyzed.

- **Amplitude-based Connection Networks.** One of the most intriguing aspects of the amplitude-based approach to evidence accumulation is that it seems to fit naturally into a connection network framework in which the nodes resemble optical lenses that transform data from signal space to knowledge space using *complex* wave patterns that can cause interference. The further possibility of layered networks, with "explainable" intermediate knowledge-space representations is interesting as well, and has exact analogies in multi-stage optical filtering.

- **Knowledge representation, propagation, and reasoning.** A potentially far-reaching aspect of the present work is the suggestion that artificially intelligent systems might perhaps benefit by using a methodology that is consistent with interactions with the physical world; that is, perhaps knowledge representation, knowledge propagation, and reasoning schemes should be phrased in terms of probability amplitudes rather than probabilities. We have no particularly strong evidence to present in favor of this proposition at this time. There is, however, a relatively simple starting point that may be used to test the utility of the formalism: one must find ways to represent knowledge in the form of solutions to linear, complex differential equations, whose solutions are identifiable as probability amplitudes for the propositions being considered. We note that just as computer simulations can carry out optical filtering calculations in abstract spaces that cannot be realized with physical optical filters, there may be extremely interesting classes of quantum-mechanics-like differential equations that are relevant to semantic reasoning; since such equations presumably have no direct analog in physical reality, they probably would not have yet been subject to extensive investigation.

# References

[1] D.H. Ballard, "Generalizing the Hough Transform to Detect Arbitrary Shapes," Pattern Recognition **13**, pp. 111–122 (1982).

[2] D.H. Ballard and D. Sabbah, "On Shapes," IJCAI-81, pp. 607–612 (1981).

[3] D.H. Ballard and C.M. Brown, *Computer Vision*, (Prentice-Hall, 1984).

[4] C.M. Brown, "Bias and Noise in the Hough Transform I: Theory," University of Rochester, Computer Science Department Report TR 105 (May, 1982); C.M. Brown and M. Curtiss, "Bias and Noise in the Hough Transform II,: Discrete Digital Results," "Bias and Noise in the Hough Transform III: Experiments with CHough," University of Rochester, Computer Science Department Reports (1982).

[5] S.R. Deans, *The Radon Transform and Some of its Applications*, (John Wiley, 1983).

[6] S.R. Deans, "Hough Transform from the Radon Transform," IEEE Trans. Pattern Anal. and Machine Perception **3**, pp. 185–188 (1981).

[7] P. Hariharan, *Optical Interferometry*, (Academic Press, 1985).

[8] R.H. Dicke and J.P. Wittke, *Introduction to Quantum Mechanics*, (Addison-Wesley, 1960).

[9] R.O. Duda and P.E. Hart, "Use of the Hough transform to detect lines and curves in pictures," CACM **15**, pp. 1–15 (1972).

[10] P. Fua and A.J. Hanson, "Objective Functions for Feature Discrimination," in Proceedings of the Eleventh International Joint Conference on Artificial Intelligence, 20–25 August 1989, Detroit, MI; pp. 1596–1602 (Morgan Kaufman, 1989); "An Optimization Framework for Feature Extraction," Machine Vision and Applic., *in press* (1990).

[11] P.V.C. Hough, *Method and Means for Recognizing Complex Patterns*, U.S. Patent 3069654 (1962).

[12] S.H. Lee, *Optical Information Processing: Fundamentals*, (Springer-Verlag, 1981).

[13] A. Papoulis, *Systems and Transforms with Applications in Optics*, (McGraw-Hill, 1968).

[14] G.O. Reynolds, J.B. DeVelis, G.B. Parrent, Jr., and B.J. Thompson, *The New Physical Optics Notebook: Tutorials in Fourier Optics*, (copublished by SPIE International Society for Optical Engineering and American Institute of Physics, 1989). See, for example, pp. 414–420.

[15] J. Rissanen, "A Universal Prior for Integers and Estimation by Minimum Description Length," The Annals of Statistics **2**, pp. 416–431 (1983).

[16] J. Rissanen, "Minimum-Description-Length Principle," in *Encyclopedia of Statistical Sciences*, **5**, pp. 523-527, (1987).

[17] D.E. Rumelhart and J.L. McClelland, *Parallel Distributed Processing*, Vols. I and II (MIT Press, 1986).

[18] G.C. Stockman and A.K. Agrawala, "Equivalence of Hough Curve Detection to Template Matching," Comm. of the ACM **20**, pp. 820–822 (1977).

[19] A. Rosenfeld and A.C. Kak, *Digital Picture Processing*, Vol. 1 and Vol. 2 (Academic Press, 1982).

[20] J. Sklansky, "On the Hough Technique for Curve Detection," IEEE TOC **C-27**, pp. 923–926 (1978).